\pdfoutput=1

\documentclass[11pt, twocolumn]{article}

\usepackage[a4paper, total={6.55in, 10in}]{geometry}
\usepackage[]{natbib}
\bibpunct[: ]{(}{)}{,}{a}{}{,}
\usepackage{authblk}

\usepackage{times}

\usepackage{multirow}

\usepackage[T1]{fontenc}

\usepackage[utf8]{inputenc}

\usepackage[dvipsnames]{xcolor}
\definecolor{darkblue}{rgb}{0, 0, 0.5}

\usepackage{hyperref}
\hypersetup{colorlinks=true, citecolor=darkblue, linkcolor=darkblue, urlcolor=darkblue}

\usepackage{graphicx}

\usepackage{enumitem}

\usepackage{xspace}
\newcommand{\taskone}{{$C_1$}\xspace}
\newcommand{\tasktwo}{{$C_2$}\xspace}
\newcommand{\vocone}{{$V_1$}\xspace}
\newcommand{\voctwo}{{$V_2$}\xspace}

\newcommand{\mix}{{\textsc{mix}}\xspace}

\title{Do Transformers know symbolic rules, \\ and would we know if they did?}
\author[1]{Tommi Buder-Gröndahl}
\author[2]{Yujia Guo}
\author[3]{N. Asokan}
\affil[1]{University of Helsinki}
\affil[2]{Aalto University}
\affil[3]{University of Waterloo}

\date{}

\begin{document}
\maketitle

\begin{abstract}
\small
To improve the explainability of leading Transformer networks used in NLP, it is important to tease apart genuine \emph{symbolic} rules from merely \emph{associative} input-output patterns.
However, we identify several inconsistencies in how ``symbolicity'' has been construed in recent NLP literature.
To mitigate this problem, we propose two criteria to be the most relevant, one pertaining to a system's internal architecture and the other to the dissociation between abstract rules and specific input identities.
From this perspective, we critically examine prior work on the symbolic capacities of Transformers, and deem the results to be fundamentally inconclusive for reasons inherent in experiment design.
We further maintain that there is no simple fix to this problem, since it arises -- to an extent -- in all end-to-end settings.
Nonetheless, we emphasize the need for more robust evaluation of whether \emph{non-symbolic} explanations exist for success in seemingly symbolic tasks.
To facilitate this, we experiment on four sequence modelling tasks on the T5 Transformer in two experiment settings: zero-shot generalization, and generalization across class-specific vocabularies flipped between the training and test set.
We observe that T5's generalization is markedly stronger in sequence-to-sequence tasks than in comparable classification tasks.
Based on this, we propose a thus far overlooked analysis, where the Transformer itself does not need to be symbolic to be \emph{part} of a symbolic architecture as the \emph{processor}, operating on the input and output as external memory components.
\end{abstract}

\section{Introduction}
\label{sec:intro}

Present-day natural language processing (NLP) is dominated by fine-tuning large pre-trained deep neural networks (DNNs) on downstream tasks.
Variants of the Transformer architecture \citep{Vaswanietal2017}
-- such as BERT \citep{Devlinetal2019}, GPT-3 \citep{Brownetal2020}, and T5 \citep{Raffeletal2020} --
have displayed impressive sequence modelling performance, including in few- and zero-shot settings \citep{Brownetal2020, Duanetal2019, Gaoetal2021, Weietal2022}.
Concurrently, model \emph{explainability} has been recognized as vital for tackling challenges that arise in standard end-to-end methodology \citep{Church2017, Church:Liberman2020, Danilevskyetal2020, Angelovetal2021}.


An essential part of model explanation is evaluating whether the model relies on \emph{symbolic rules} beyond merely \emph{associative} patterns \citep{Marcus2018, Marcus2020, kassner2020pretrained, Hupkes2020}.
However, we note major disparities in what has been considered ``symbolic'' across contemporary NLP literature.
As reviewed in Section \ref{sec:background}, proposed criteria encompass e.g. out-of-distribution generalization \citep{Marcus2020}, discreteness \citep{Cartuyvelsetal2021}, semantics \citep{Santoroetal2022}, the type of data \citep{Lample:Charton2020}, and the use of information not explicitly seen during training \citep{kassner2020pretrained}.
Problematically, these do not converge to a unified conception of symbolicity.

To unravel the situation, in Section \ref{sec:problem-statement} we identify two notions of symbolicity that we consider the most apt for evaluating DNNs.
The first builds on \citet{Turing1937} in taking symbolic computation to involve read/write operations on stored representations (e.g. \citealt{Gallistel:King2010}).
The second treats a rule as symbolic if it is \emph{abstract} in making no reference to specific input identities, and thereby bolsters \emph{systematic generalization} \citep{Fodor:Pylyshyn1988, Aizawa2003, McLaughlin2009}.
The criteria are conjoined in prototypical symbolic systems, but the use of abstract rules is not guaranteed by the use of a read/write memory.

Given the two criteria, we critically examine prior results on the performance of Transformers in putatively symbolic tasks.
We deem these to be fundamentally inconclusive due to an inherent ambiguity in experiment design.
The dilemma boils down to the \emph{the unfalsifiability of non-symbolic explanations} of model performance.

While we do not foresee simple fixes to the predicament, we stress the importance of considering possible non-symbolic solutions to seemingly ``symbolic'' tasks: it should be established whether observed performance could be attained in the absence of genuine symbolic rules.
One way to advance this is line of research is to examine multiple comparable experiment settings with minor modifications to pinpoint relevant loci of variation.

As case studies, Section \ref{sec:methodology} presents four sequence-to-sequence (seq2seq) or sequence classification tasks applied to T5 \citep{Raffeletal2020}: copying/reversal, copy/reverse detection, palindrome detection, and repetition detection.
We further distinguish between two experiment settings, both concerning model generalization beyond the fine-tuning vocabulary distribution.
In the \emph{zero-shot} setting, the model is fine-tuned on one vocabulary and tested on another.
In the \emph{vocabulary flip} setting, the model is fine-tuned with target class -specific vocabularies that are flipped in the test set.

We find three main divergencies in T5's performance between the tasks, covered in Section \ref{sec:results}:

\begin{enumerate}[nosep]
\item Zero-shot performance significantly exceeds vocabulary flip performance across all tasks.
\item Generalization is substantially stronger in seq2seq tasks than in classification tasks.
\item When small ($1\%-10\%$) ratios of datapoints from flipped vocabularies are added in fine-tuning, T5 continues to fail at repetition detection despite succeeding at other tasks.
\end{enumerate}

As discussed in Section \ref{sec:discussion}, these differences in task performance shed light on the presence of symbolic rules.
In particular, T5's marked difficulty to learn repetition detection indicates an underlying obstacle to internalizing genuinely abstract symbolic rules that shun associative heuristic solutions.

Furthermore, motivated by the initially puzzling contrast between the seq2seq and classification tasks, we propose a novel hypothesis concerning the symbolic capacities of Transformers.
The architectural conception of symbolicity relies on the \emph{memory-processor distinction}.
Crucially, it is agnostic about the location of these components in the implementing system.
This brings about a third possibility between treating Transformers as either fully associative or fully symbolic: they can function as the \emph{processor} component in a larger system with the input and output as external ``memory'' components.
Specifically, we suggest that attention together with positional encoding can implement \emph{location-based memory} for storing abstract rules.
However, its use is severely limited in classification, where all input information must be processed model-internally in a single encoding step.
This hypothesis thus predicts the observed divergence between seq2seq and classification performance in learning abstract rules.

In summary: even if the model alone is not symbolic, the full input-model-output pipeline might be.
This indicates that recent successes with augmenting Transformer input \citep{Borgeaudetal2021, Nakanoetal2021} could have a fundamental computational basis: externalizing memory facilitates the learning of abstract rules.

The paper's contributions are outlined below:

\begin{itemize}[nosep]
\item We show the lack of a unified conception of ``symbolic'' in contemporary NLP (Section \ref{sec:background}).
\item We identify a deep-seated challenge in interpreting prior results on symbolic learning by DNNs, and propose the methodology of scrutinizing loci of variation across comparable \emph{prima facie} symbolic tasks (Section \ref{sec:problem-statement}).
\item Experimenting on T5's vocabulary generalization abilities across four tasks (Section \ref{sec:methodology}), we highlight the following patterns (Section \ref{sec:results}):
\begin{itemize}[nosep]
    \item T5 struggles to learn abstract rules when associative heuristics are unavailable.
    \item Generalization is significantly better in seq2seq tasks than in classification tasks.
\end{itemize}
\item Instead of treating Transformers as fully symbolic or non-symbolic, we propose a third alternative where they can function as the processor in a larger symbolic architecture with external memory components (Section \ref{sec:discussion}).
\end{itemize}

\section{Background}
\label{sec:background}

\citet{Fodor:Pylyshyn1988} originally raised the question of whether connectionist models display \emph{systematicity}, where certain computational capacities are reliably linked with others. For example, a systematic cognitive system able to understand \emph{John saw Mary} could also understand \emph{Mary saw John}. Controversially, \citeauthor{Fodor:Pylyshyn1988} further asserted that systematicity requires a Turing architecture that encodes symbolic rules, which at least simple connectionist systems (e.g. classical perceptrons) lack. The matter has remained a topic of long-lasting debate without a general consensus \citep{Aizawa2003, McLaughlin2009, Kiefer2019}.
Similar fundamental questions have arisen for state-of-the-art DNNs such as LSTMs \citep{Hochreiter:Schmidhuber1997}, GRUs \citep{Choetal2014}, and Transformers \citep{Vaswanietal2017}.

DNN-based NLP has largely followed an end-to-end approach that focuses on improving model performance rather than understanding it theoretically \citep{Church2017, Church:Liberman2020}. However, increasing emphasis has recently been placed on \emph{model explainability} \citep{Danilevskyetal2020, Angelovetal2021}.
This development is manifested by the common use of human-readable symbolic rules for analyzing the results of procedures like attention visualization \citep{Thorneetal2019, Vig2019} or structural probing \citep{Hewitt:Manning2019, Chenetal2021}.
For example, many studies have interpreted BERT as constructing classical linguistic representations such as syntax trees, dependency graphs, or semantic roles \citep{Jawaharetal2019, Kovalevaetal2019, Tenneyetal2019, Manningetal2020}.
These are paradigm examples of symbolic representations.


Symbolic interpretations of DNNs have also been challenged.
\citet{Lake:Baroni2018} illustrate the difficulties of LSTMs in tasks that require systematic inference.
\citet{Goodwinetal2020} argue that LSTMs and GRUs encode input tokens in context-sensitive ways that discourage systematic generalization.
Similar results are corroborated on Transformers by \citet{Talmoretal2020}.
\citet{Hupkes2020} evaluate LSTMs and Transformers on compositional inference, and deem that none of their models reliably exhibit it.
\citet{kassner2020pretrained} evaluate BERT's abilities on six symbolic reasoning tasks, with only partial success.
 
 DNNs have a general tendency to rely on surface heuristics
 \citep{Gururanganetal2018, McCoyetal2019, Mickusetal2019}, which can lead to biases \citep{Kuritaetal2019, Nadeemetal2021} and susceptibility to adversarial examples \citep{Lietal2020}.
Such considerations indicate that models often prefer associative pattern matching to abstract rules.
Still, this does not mean that they lack the capacity for understanding abstract rules \emph{in principle}.

However, a deep-seated problem underlying prior research is that the notion of ``symbolic'' is rarely made explicit.
Often the word is simply used without a proper explanation.\footnote{For instance, \citet{Hoehndorf:Queralt-Rosinach2017} state that symbolic systems ``represent things (...) through physical \emph{symbols}, combine \emph{symbols} into \emph{symbol} expressions, and manipulate \emph{symbols} and \emph{symbol} expressions'' (p. 27; our emphases).}
Furthermore, suggested criteria for symbolicity vary significantly.
Below, we briefly review seven characteristics commonly assigned to symbols.

\noindent{\textbf{Human-readability.}}
Symbolic representations are often (at least potentially) understandable to a human interpreter. However, this is neither necessary nor sufficient for symbolicity, as associative models can also allow feature interpretation, and symbolic computation could exceed human comprehension. 

\noindent{\textbf{Manual programming.}}
A well-known drawback of rule-based AI is the need for human labor in programming \citep{Bengioetal2021}.
However, this is not constitutive of symbolicity, which concerns the nature of the computation rather than its origin.

\noindent{\textbf{Semantics.}}
In the field of semiotics, symbols are taken to be signs that bear an arbitrary relation to their referents, based on convention rather than on resemblance (as in iconic signs) or a factual relation (as in indexical signs) \citep{Peirce1868}.
While this conception has typically been distinct from the computational notion of ``symbol'', it has recently been raised in NLP as well \citep{Santoroetal2022}.
Alternatively, the word ``symbol'' can be used of anything that has a semantic interpretation, basically assimilating to the semiotic notion of sign.
This reflects an influential view in classical cognitive science, according to which computation requires representational content \citep{Fodor81}.
However, the necessity of semantic interpretation for computational symbols has also been challenged \citep{Pylyshyn1984, Egan2010, Piccinini15}.
They indeed seem separable especially in an AI context: for example, our experiments use artificial tokens with no semantic interpretation attached (Section \ref{sec:methodology}).
The relationship between formal computation and semantic content deserves more scrutiny in NLP (cf. \citealt{Bender:Koller2020}).

\noindent{\textbf{Discreteness.}}
Symbols are sometimes assimilated to discrete/digital units or structures, set against the contiguous/analog representations used by DNNs (e.g. \citealt{Bengioetal2021, Cartuyvelsetal2021}).
However, this is problematic in both directions. Discrete representations and operations can also be used in non-symbolic systems, such as classical perceptrons \citep{Rosenblatt1958}. Moreover, symbolic computation can be either digital or analog \citep{Gallistel:King2010, Piccinini15}.

\noindent{\textbf{Symbolic vs. numerical mathematics.}}
Mathematical expressions are symbolic if they contain variables instead of specific numerical values.
\citet{Lample:Charton2020} train a Transformer network on data representing symbolic equations, and treat success in this task as an indicator of symbolic computation by the model. A potential problem here is that the symbols are part of the input and output, and it is not evident whether mapping these to each other requires \emph{model-internal} symbols.
Furthermore, symbolic mathematics is clearly insufficient to define symbolic computation more broadly, which can apply beyond mathematical expressions.

\noindent{\textbf{Model generalization.}}
The end-to-end focus of DNN applications has resulted in symbolicity being given mostly operational definitions.
\citet{kassner2020pretrained} consider the criterion for symbolic reasoning to be the ability to ``infer knowledge not seen explicitly during pretraining'' (p. 552).
However, it is unclear why this could never arise in a non-symbolic model.
A more specific idea is that the models should generalize beyond the training \emph{distribution} (e.g. \citealt{Marcus2018, Marcus2020}).
This, too, is inadequate without further elaboration, as non-symbolic associative processes could also generalize across certain aspects of training and test distributions.
For example, \citet{Bengio2019} specifically advocates the goal of increasing model generalizability \emph{without} introducing symbolic computation.

\noindent{\textbf{Memory-processor distinction.}}
Finally, a classical \emph{architectural} criterion for symbolic systems is the presence of a \emph{read-write memory} which is accessed by a separate \emph{processor} that enacts read and write operations (e.g. \citealt{Gallistel:King2010}). Here, symbols are units stored in the memory and manipulated by the processor.
In Turing machines \citep{Turing1937} the memory-processor distinction corresponds to the division between the tape and the read/write head, and in von Neumann architectures \citep{vonNeumann1945} to the division between the memory unit and the processor.

Most of the proposed criteria covered above are neither necessary nor sufficient for symbolicity, as they can be present in non-symbolic systems and absent from symbolic systems.
At best, they pinpoint prototypical properties of systems/tasks commonly called ``symbolic'' in the literature, constituting \emph{family resemblance} \citep{Wittgenstein1953, Rosch1975} rather than a definition.
An important exception to this is the memory-processor distinction, which is a clearly delineated property of the implementing system's architecture.
In Section \ref{sec:problem-statement}, we maintain that the most relevant notions of symbolicity for evaluating DNNs are this architectural conception on the one hand, and the learning of \emph{abstract rules} on the other hand.
\section{Problem statement}
\label{sec:problem-statement}


Based on the architectural conception of symbolic computation, \citet{Gallistel:King2010} contend that neural networks are incapable of implementing it due to their \emph{holistic} nature.
Their internal states are highly entangled with respect to the information they hold: the information given by one part of the network typically depends on many (possibly all) other parts. Changes to the network thus have global consequences, which prevents their internal division into distinct memory and processor components.
A related idea is expressed by \citet[p. 143]{Piccinini15}, who suggests that neural networks are \emph{computationally primitive} in the same way as e.g. logic gates: while they can participate in computation, their own internal operations cannot be computationally decomposed.\footnote{Our proposal in Section \ref{sec:discussion} bears important similarity to this in distinguishing between model-internal symbolicity and the symbolicity of the full input-model-output pipeline.}

Nevertheless, such theoretical considerations alone are insufficient to rule out the possibility of a model-internal memory-processor distinction emerging in a DNN; especially given the success of Transformers in ostensibly ``symbolic'' tasks such as mathematical reasoning \citep{Lample:Charton2020} or linguistic inference \citep{Tafjordetal2021}.
Experimental evaluation is also needed.

Symbolicity should further be connected to the internalization of \emph{abstract rules}.
This is the focus of much prior work (see Section \ref{sec:background}), and ignoring it would significantly decrease the relevance of the topic for NLP.
For present purposes, we consider a rule to be abstract if it \emph{does not refer to specific input identities}.
The implementation of any rule is always vocabulary-bound in the sense that concrete systems only recognize finite vocabularies. However, there is a crucial difference between this and the rule itself referring to specific inputs.
As a toy example, the rule `map $a$ to $a$ and $b$ to $b$' is input-specific; whereas the rule `map any input to itself' is not -- even if the system only recognized the input vocabulary \mbox{$\{a, b\}$}.

Abstract rules are especially manifested in \emph{variable binding}, which is a crucial aspect of symbolic computation \citep{Marcus2001, Marcus2018, Marcus2020}.
This requires the system to implement  \emph{binding} relations that link \emph{variables} to \emph{instances}, and computational operations involving variables rather than only the instances.
Both the variable and instance are symbols held in memory.
In von Neumann architectures, variable binding is implemented via \emph{location-based addressing} and \emph{pointer architectures}.\footnote{Similar ideas have also been adopted in cognitive psychology \citep{Green:Quilty-Dunn2017, Quilty-Dunn2020} and neuroscience (e.g. \citealt{Krieteetal2013}).}
Crucially, the memory-processor distinction allows the \emph{indefinite maintenance} of the binding relation across changes to the rest of the system.

However, it is important to note that a memory-processor distinction does not as such require the system to learn abstract rules.
Conversely, it is not obvious why an associative system could not learn anything ``abstract'' in some sense of the term.
Hence, the link between symbolicity and abstract rules is less apparent than might initially seem.

We draw the connection by elaborating on \citeauthor{Gallistel:King2010}'s (\citeyear{Gallistel:King2010}) observation discussed above: if a system is holistic in not admitting separation between its internal components, it also will not admit a distinction between the representations of abstract rules and input-specific information. Therefore, a genuinely abstract (i.e. input-independent) rule can only be represented by a system with an internal separation between a component for representing the rule in memory, and another component for representing input-specific information that interacts with the memorized abstract rule in determined ways without becoming fully ``entangled'' with it.
Without a memory-processor distinction such entanglement would be unavoidable, since the input could only impact the system as a whole.

Despite this theoretical foundation for linking abstract rule learning to architectural symbolicity, we note a major dilemma in interpreting prior experimental results: the lack of baseline information on how a \emph{non-symbolic} system would perform. This is illustrated by the quote below:
 
 \begin{quote}
     ``It has been shown that such a language model [as BERT] \emph{contains} certain degrees of syntactic \citep{Goldberg2019}, semantic \citep{Clarketal2019}, common-sense \citep{Cuietal2020} and logical reasoning \citep{Clarketal2020} \emph{knowledge}.'' \\ (\citealt[p. 13392]{Liuetal2021}; our emphases; references reformatted by us)
 \end{quote}
 
The quote is ambiguous. It is unproblematic if ``knowledge'' is understood in the \emph{procedural} sense of ``knowing how'' rather than the \emph{propositional} sense of ``knowing that'' \citep{Ryle49}: the model has e.g. ``logical knowledge'' if it performs sufficiently well in end-to-end tasks that can be characterized as ``logical reasoning''. In contrast, the shift from procedural to propositional knowledge requires the crucial additional assumption that procedural knowledge could \emph{only} be achieved via propositional knowledge.
Applied to AI, the inference from task performance to the internalization of symbolic rules relies on the assumption that such internalization is necessary for obtaining the performance.
Summarized below, the argument from premises P1--P2 to the conclusion is only valid with the additional premise P3: \\

\begin{itemize}[nosep]
    \item[P1:] Task T can be described by rule R
    \item[P2:] Model M succeeds in T
    \item[\textbf{P3:}] \textbf{M would fail in T without internalizing R}
    \item[$\Rightarrow$] M has internalized R \\
\end{itemize}

Interpreting prior studies (see Section \ref{sec:background}) as favoring(/opposing) symbolic computation in DNNs would require P3 as a background assumption, but baseline information in its support is lacking.
Instead, model performance in tasks that can be \emph{described} by a rule has readily been treated as direct evidence for(/against) the model having \emph{internalized} the rule.
This is insufficient in the absence of knowledge on how well the model could perform in the task \emph{without} having internalized the rule.
 
Non-symbolic associative models can mimic abstract rules by responding in a similar way to different inputs, which can seem like abstracting away from the input. In contemporary DNNs the most apparent basis of this is \emph{embedding similarity} between input tokens, which has been suggested to underlie e.g. generalization between languages in multilingual BERT \citep{Ceoetal2020}.
Unlike with genuinely abstract rules, here model performance is still embedding-specific: generalization arises from the similar embeddings of multiple tokens rather than abstract (input-independent) rules.
 
Therefore, to evaluate whether DNNs genuinely learn abstract rules, there should be a robust way to rule out model success via associative means based on e.g. embedding similarities. In end-to-end settings, there is currently no reliable method to achieve this.
The dilemma is thus not a technical flaw in prior work, but instead reflects a fundamental challenge in model interpretation.

Despite the lack of simple solutions, we emphasize the need for more careful consideration of possible associative means for tackling seemingly ``symbolic'' tasks.
Even if genuine proofs of either symbolic or non-symbolic interpretations of DNNs were unavailable, such evaluation is crucial especially for assessing the credibility of symbolic interpretations.
In particular, we propose comparing multiple \emph{prima facie} symbolic tasks across similar experiment settings to discover possible points of divergence in model performance, and creating a comprehensive taxonomy of the results to facilitate large-scale comparison between symbolic and associative interpretations.

\section{Methodology}
\label{sec:methodology}

As case studies, we evaluate the T5 Transformer \citep{Raffeletal2020} on four simple tasks that require generalizing rules across two disjoint subsets of the vocabulary: \vocone and \voctwo.
Our source code for reproducing the experiments will be made available on GitHub, and is provided as supplementary material along with raw data.

\noindent{\textbf{Tasks.}}
We experimented on a seq2seq task with two variations, a corresponding sequence pair classification task, and two sequence classification tasks, summarized below and in Table \ref{tab:tasks}:\footnote{The task formulation is inspired by related experimental work conducted on MLPs in the 1990s (see   \citealt{Marcus2001}).} \\

\begin{itemize}[nosep]
    \item \textbf{Copy/reverse:} how to either repeat or reverse the input sequence?\footnote{To avoid ambiguity, we discarded palindromes in the fine-tuning and test data both here and in copy/reverse detection.}
    \item \textbf{Copy/reverse detection:} are two sequences copies or reversals of each other?
    \item \textbf{Palindrome detection:} is a sequence identical with its reversal?
    \item \textbf{Repetition detection:} does a sequence contain more than one instance of any token? \\
\end{itemize}

\noindent{\textbf{Data.}}
We used individual lowercase letters as input tokens, divided $50\%-50\%$ into \vocone and \voctwo.
In the seq2seq task we distinguished task classes by the prefix ``copy'' or ``reverse''.
The simplicity of our vocabulary choice aims at minimizing collateral influence due to T5's pre-training.

We adopted two experiment settings, titled \emph{zero-shot} and \emph{vocabulary flip}. These differ in how \vocone and \voctwo relate in the fine-tuning and test data.

\noindent{\textbf{Zero-shot.}}
In this setting, we fine-tuned T5 on \vocone and tested it on \voctwo.
This is made possible by both vocabularies being recognized by the pre-trained T5 tokenizer prior to fine-tuning.

\noindent{\textbf{Vocabulary flip.}}
Here, we split inputs into two \emph{task classes}: \taskone and \tasktwo. In classification these correspond to distinct target classes, and in seq2seq they are separated by task prefixes.
We used \vocone for \taskone and \voctwo for \tasktwo during fine-tuning.
The fine-tuned model was then tested with flipped vocabularies: \voctwo for \taskone and \vocone for \tasktwo.
We additionally introduced random mixing of vocabularies and task classes in the fine-tuning data, regulated by a \emph{mix ratio} \mix.
This is the probability of switching the vocabulary-task pairing in any datapoint.
If $\mix=0$, no such switching occurs.
A low but non-zero \mix maintains the vocabulary bias but gives the model ``hints'' that the rule should be vocabulary-general.\footnote{Mixing was not applied in the test set. Introducing test-like datapoints to the training set was also used by \citet{Lake:Baroni2018} for LSTMs in systematic inference tasks. Here, low mixing did not significantly improve model performance.}

\noindent{\textbf{Hyperparameters.}}
We fine-tuned the \emph{t5-base} model,\footnote{\url{https://huggingface.co/t5-base}} using the vocabulary size of $10$ for both \vocone and \voctwo, $10000$ as both the fine-tuning and test set size with a $80\%$ evaluation split, and the batch size of $16$.
Training for $20$ epochs, we applied the model checkpoint with the lowest evaluation loss to the test set.
Further details on hyperparameters and implementation are provided in Appendix \ref{sec:appendix-hyperparameters}.

\begin{table}[t]
  \begin{center}
     \begin{tabular}{|c|c|c|c|} \hline
    \multirow{2}{*}{\textbf{Task}} & \multirow{2}{*}{\shortstack{\textbf{Task} \\ \textbf{class}}} & \multicolumn{2}{c|}{\textbf{Example}} \\ \cline{3-4}
    && \textbf{input} & \textbf{output} \\ \hline

    copy/reverse & \taskone & $a$ $b$ & $a$ $b$ \\ (seq2seq) & \tasktwo & $a$ $b$ & $b$ $a$ \\ \hline
    
    copy/reverse & \taskone & $a$ $b$ </$s$> $a$ $b$ & $1$ \\ detection & \tasktwo & $a$ $b$ </$s$> $b$ $a$ & $0$ \\ \hline

    palindrome & \taskone & $a$ $b$ $a$ & $1$ \\ detection & \tasktwo & $a$ $b$ $b$ & $0$ \\ \hline

    repetition & \taskone & $a$ $b$ $a$ & $1$ \\ detection & \tasktwo & $a$ $b$ $c$ & $0$ \\ \hline
   
    \end{tabular}
    \caption{\small Sequence modelling tasks experimented on.}
    \label{tab:tasks}
  \end{center}
\end{table}
\section{Results}
\label{sec:results}

We cover T5's performance in zero-shot settings (Section \ref{sec:results-zero-shot}) and vocabulary flip settings with $\mix \in \{0\%, 1\%, 10\%\}$ (Section \ref{sec:results-vocabulary-flip}).
Training and evaluation accuracies were consistently high ($0.91-1.00$; see Appendix \ref{sec:appendix-results}), and Table \ref{tab:results} presents test set accuracies for each task class.
We highlight the \emph{discrepancy between seq2seq and classification tasks}, especially in the vocabulary flip setting.
(All numbers are rounded to two decimal places.)

\begin{table*}[t]
  \begin{center}
    \begin{tabular}{|c|c|c|c|c|c|} \hline
    \multirow{2}{*}{\textbf{Task}} & \multirow{2}{*}{\textbf{Task class}} & \multirow{2}{*}{\textbf{Zero-shot}} & \multicolumn{3}{c|}{\textbf{Vocabulary flip}} \\ \cline{4-6}
    &&& \mix$=0\%$ & \mix$=1\%$ & \mix$=10\%$ \\ \hline
    
    copy/reverse & copy & $\textcolor{Green}{1.00}$ &
    $\mathbf{0.75}$ &
    $\textcolor{Green}{1.00}$ &
    $\textcolor{Green}{1.00}$ \\
    (seq2seq) & reverse &
    $\textcolor{Green}{0.97}$ &
    $\mathbf{0.72}$ &
    $\textcolor{Green}{0.99}$ &
    $\textcolor{Green}{0.99}$ \\ \hline
    
    copy/reverse & copy &
    $\textcolor{Green}{1.00}$ &
    $\textcolor{red}{0.00}$ &
    $\textcolor{Green}{1.00}$ &
    $\textcolor{Green}{1.00}$ \\
    detection & reverse &
    $\mathbf{0.88}$ &
    $\textcolor{red}{0.00}$ &
    $\textcolor{Green}{0.94}$ &
    $\textcolor{Green}{1.00}$ \\ \hline
    
    palindrome & palindrome &
    $\textcolor{Green}{1.00}$ &
    $\textcolor{red}{0.07}$ &
    $\textcolor{red}{0.09}$ &
    $\textcolor{Green}{0.90}$ \\
    detection & non-palindrome &
    $\textcolor{Green}{0.90}$ &
    $\textcolor{red}{0.00}$ &
    $\textcolor{red}{0.02}$ &
    $\textcolor{Green}{0.91}$ \\ \hline
    
    repetition & repetition &
    $\textcolor{Green}{0.96}$ &
    $\textcolor{red}{0.08}$ &
    0.27 &
    $\mathbf{0.30}$ \\
    detection & no repetition &
    $\textcolor{Green}{1.00}$ &
    $\textcolor{red}{0.10}$ &
    0.12 &
    $\mathbf{\textcolor{red}{0.10}}$ \\ \hline
   
    \end{tabular}
    \caption{Test set accuracy per task class in zero-shot and vocabulary flip settings with $\mix \in \{0\%, 1\%, 10\%\}$ \\ (\textcolor{Green}{green:} $\geq 0.90$;
    \textcolor{red}{red:} $\leq 0.10$;
    \textbf{bold:} deviation from the general pattern).}
    \label{tab:results}
  \end{center}
\end{table*}

\subsection{Zero-shot}
\label{sec:results-zero-shot}

T5 succeeded to a significant degree in all zero-shot settings, displaying strong generalization from \vocone to \voctwo despite never seeing \voctwo during fine-tuning. 
Accuracy was the lowest for the reverse class in copy/reverse detection: only $0.88$ in comparison to $1.00$ for the copy class in the same task, and $0.97$ for the reverse variant of the seq2seq task.
Thus, while T5 learned both to reverse a string and detect if two strings are each others' reversals, its performance on the latter was markedly lower.

\subsection{Vocabulary flip}
\label{sec:results-vocabulary-flip}

All three classification tasks completely failed with \mbox{$\mix=0$}, only reaching accuracies in the range $0.00-0.10$. Increasing \mix to $1\%$ induced success in copy/reverse detection ($0.94-1.00$), and \mbox{$\mix=10\%$} in palindrome detection ($0.90-0.91$).
In contrast, repetition detection never exceeded $0.30$ accuracy in either class,
performing clearly below the chance level ($0.50$) even with $\mix=10\%$.

As opposed to classification, the copy/reverse seq2seq task obtained $0.72-0.75$ accuracy already with \mbox{$\mix=0$}, increased to $1.00$ with a higher \mix.
Thus, while generalization from \vocone to \voctwo was nonexistent in the classification tasks with \mbox{$\mix=0$}, in seq2seq T5 learned to apply the correct rule the clear majority of time even here.

\subsection{Summary}
\label{sec:results-summary}

Zero-shot results establish T5's ability to robustly extend across vocabularies in \emph{prima facie} symbolic tasks.
At the same time, the failure of repetition detection in the vocabulary flip setting illustrates that T5 still struggles with vocabulary bias in learning abstract rules, even when given explicit counter-evidence via \mix.
Vocabulary generalization was systematically stronger in seq2seq than in classification, especially in the vocabulary flip setting. 
\section{Discussion}
\label{sec:discussion}



It is possible to treat T5's zero-shot success as corroborating its symbolic interpretation.
According to this analysis, the model has internalized abstract rules (e.g. `copy/reverse the input'), stored using some kinds of placeholder symbols for input variables (e.g. `first/second/... element').
In contrast, a rival interpretation is that the results illustrate the aptitude of \emph{associative} learning, and thus actually \emph{cast doubt on the necessity of symbolicity} for achieving vocabulary generalization.

The difficulty of deciding between these contenders exemplifies the dilemma discussed in Section \ref{sec:problem-statement}: the symbolic interpretation effectively relies on rejecting non-symbolic alternatives \emph{a priori}.
However, it is not ruled out that the detection of two identical input tokens could be based on e.g. the comparison between their corresponding embedding components.\footnote{\label{fn:embedding-component-comparison} As a simple example, vectors \mbox{$v_1 = (0, 1)$} and \mbox{$v_2 = (1, 0)$} share the range of component values. A model that compared pairs of corresponding components between two vectors would thus compare $0$ to $0$ and $1$ to $1$ \emph{both} when comparing $v_1$ to itself and when comparing $v_2$ to itself. If the decision of whether two tokens are identical is based on the comparison of their corresponding embedding components, this could conceivably facilitate zero-shot success in detecting identical embeddings in different input positions.}
Such non-symbolic analyses of our zero-shot results remain hypothetical at present, but provide an important venue for future research.


The lack of vocabulary flip generalization in classification tasks with $\mix=0\%$ is not particularly surprising as such:
T5 gets caught on the vocabulary bias as a simple heuristic instead of learning the abstract rule.
In contrast, the two main task discrepancies are less straight-forward to account for: \mbox{(i) the} particular failure at repetition detection even with $\mix=10\%$; and \mbox{(ii) the} success of seq2seq compared to classification.
The former indicates a resistance to learning the abstract rule \emph{even when explicit counter-evidence is provided} against the vocabulary heuristic. The difference between the three classification tasks is plausibly explained by task difficulty, allowing an increased \mix to induce learning the \voctwo variant for the two simpler tasks.\footnote{Both copy/reverse detection and palindrome detection rely on \emph{position-specific} comparison between tokens; whereas repetition detection requires finding two instances in \emph{any positions}. The former are thus expected to be easier to learn via positional attention. As of yet, we have no explanation of the difference between copy/reverse detection and palindrome detection with $\mix=1\%$, and further research is required to evaluate whether similar effects reliably recur.}


Crucially, however, vocabulary generalization \emph{is} possible even with $\mix=0$, as illustrated by the copying/reversal seq2seq task.
To explain this, we propose that here the model can use input and output as \emph{external memories}. Based on the architectural conception, this yields a \emph{symbolic} interpretation of the input-Transformer-output pipeline.

Transformers apply \emph{multi-head attention} to input positions for producing contextual embeddings of input tokens \citep{Vaswanietal2017}.
At each copy/reverse step, the model first needs to attend to the input position of the relevant token, and then replicate this token in the output.
Attention can be learned via the task marker token: the encoder should begin with the sequence-initial position and increase the position by one at each step when copying; and the converse when reversing.
Token replication is a one-to-one mapping task learnt separately for each token. Zero-shot copy/reverse success further indicates that T5's pre-training already facilitates token replication
(possibly due to the paraphrasing pre-training task of \emph{t5-base}).

\begin{figure}[t]
\includegraphics[width=8.6cm]{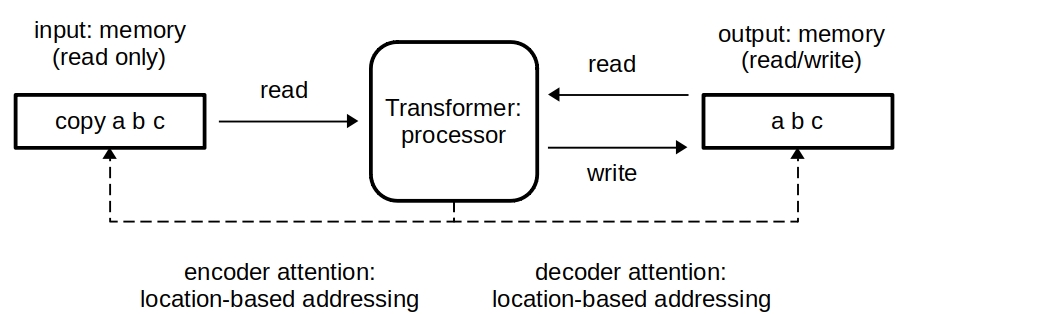}
\caption{Architecturally symbolic interpretation of the full input-Transformer-output pipeline.}
\label{fig:symbolic-pipeline}
\end{figure}

A distinction thus arises between \mbox{(i) attending} to a certain input position; and \mbox{(ii) replicating} the token in the attended position.
The first part bears important resemblance to \emph{location-based memory addressing}, where a memory address is read \emph{irrespective of its content}. Here, the input position corresponds to a memory location, the input token to the content of the location, and positional attention to the \emph{read} operation. In token replication,
decoding the token corresponds to a \emph{write} operation on the output as an external memory component, and 
attending to prior decoder output to another \emph{read} operation akin to reading the input.
In other words, the Transformer functions as a \emph{processor} within a larger symbolic architecture containing the input and output as external read/write memories.
Figure \ref{fig:symbolic-pipeline} presents the overview of this analysis.

Unlike seq2seq, classification cannot rely on attention to different input positions at each encoding step, since there is \emph{only one} step.
Multi-step computation (such as repetition detection) would thus need to take place \emph{model-internally}. This requires the model to rely on its own memory resources without being able to ``externalize'' location-addressable memory to the input/output.

Significant improvement in Transformer performance has recently been achieved with augmented input \citep{Borgeaudetal2021, Nakanoetal2021}. Our analysis suggests a computational explanation of these findings: encouraging the use of external memory allows the model to focus fully on its tasks as the processor, which facilitates the architectural symbolicity of the full pipeline.
This is further corroborated by theoretical work that highlights the importance of external input for reaching Turing-completeness in Transformers \citep{Perezetal2021}.
\section{Conclusions and future work}
\label{sec:conclusions}

The question of whether Transformers internalize symbolic rules remains open, and will likely require algorithmic analysis beyond experimental work.
The observed discrepancy between zero-shot and vocabulary flip generalization also invites further investigation.
We stress the importance of evaluating (as of yet unfalsified) \emph{non-symbolic} explanations before drawing conclusions on genuine model-internal symbolicity.

In contrast, we propose that the full input-Transformer-output pipeline satisfies the architectural requirement of symbolicity (memory-processor distinction): positional attention can implement location-based memory addressing with the input and output as external read/write memories.
The potential of using Transformers for symbolic computation thus crucially depends on the \emph{range of available input and computational steps}.
\section*{Limitations}
\label{sec:limitations}

In order to facilitate the interpretability of results, our experimental setup is -- by design -- simple and formulated in artificial settings.
This allows us to focus on the main research questions by minimizing confounding effects that may arise in more complex (and hence more realistic) NLP settings.
Our vocabularies are artificial and maximally simple (single characters), and the model architecture is restricted to T5.
Future work should examine the performance of tokens with more prevalent linguistic functions, as well as compare different linguistic classes (e.g. content vs. function words) and different languages in multi-lingual models.
Our classification tasks are restricted to binary targets, and similar experiments could also be conducted in \emph{multi-class} as well as \emph{multi-label} settings.
Experiments should also be expanded beyond T5 to cover more Transformer variants, evaluate the impact of model hyperparameters, and compare Transformers with a wider range of DNN architectures.

While we contend that non-symbolic associative analyses of zero-shot success on identical token detection have not been demonstrably falsified (Section \ref{sec:discussion}, Footnote \ref{fn:embedding-component-comparison}), this does not yet show such analyses to be correct.
Additional research is needed to establish whether non-symbolic explanations of \emph{prima facie} symbolic model performance are not only possible in principle but also plausible in practice.

\section*{Ethics statement}
\label{sec:ethics}

This project involved no experiments on humans or non-human animals, and no privacy-sensitive data or offensive material was used or produced. The source code was built using open-source libraries, and a complete pipeline for replicating all experiments will be made available as open-source.


\bibliography{refs}

\newpage
\appendix
\section{Experiment hyperparameters}
\label{sec:appendix-hyperparameters}

We used \emph{Pytorch}\footnote{\url{https://pytorch.org/}} (1.7.0) as the deep learning framework, \emph{simpletransformers}\footnote{\url{https://simpletransformers.ai/}} (0.63.4) for obtaining T5, and \emph{t5-base}\footnote{\url{https://huggingface.co/t5-base}} as the pre-trained model for fine-tuning. We conducted GPU computation on the Docs CSC computing platform with NVIDIA V100.
The average training time for $20$ epochs was $84$ minutes.

Table \ref{tab:hyperparameters} lists the hyperparameters.

\begin{table}[h!]
  \begin{center}
    \begin{tabular}{|c|c|c|c|} \hline

    \multirow{9}{*}{\shortstack{\textbf{Pre-} \\ \textbf{trained}}} &

    \textbf{model} & \multicolumn{2}{c|}{\emph{t5-base}} \\ \cline{2-4}
    & \textbf{parameters} & \multicolumn{2}{c|}{$220$M} \\ \cline{2-4}
    & \multirow{2}{*}{\shortstack{\textbf{vocabulary} \\ \textbf{size}}} & \multicolumn{2}{c|}{\multirow{2}{*}{$32 000$}} \\ && \multicolumn{2}{c|}{} \\ \cline{2-4}
    & \multirow{2}{*}{\shortstack{\textbf{attention} \\ \textbf{heads}}} & \multicolumn{2}{c|}{\multirow{2}{*}{$12$}} \\ && \multicolumn{2}{c|}{} \\ \cline{2-4}
    & $\mathbf{d_{model}}$ & \multicolumn{2}{c|}{$768$} \\ \cline{2-4}
    & $\mathbf{d_{kv}}$ & \multicolumn{2}{c|}{$64$} \\ \cline{2-4}
    & $\mathbf{d_{ff}}$ & \multicolumn{2}{c|}{$3072$} \\ \hline

    \multirow{9}{*}{\shortstack{\textbf{Fine-} \\ \textbf{tuning}}} &

    $\mathbf{V_1}$ & \multicolumn{2}{c|}{abcdefghij} \\ \cline{2-4}
    & $\mathbf{V_2}$ & \multicolumn{2}{c|}{klmnopqrst} \\ \cline{2-4}

    & \multirow{3}{*}{\shortstack{\textbf{dataset} \\ \textbf{size}}} & train & $8000$ \\
    && eval & $2000$ \\
    && test & $10000$ \\ \cline{2-4}
    
    & \multirow{2}{*}{\shortstack{\textbf{datapoint} \\ \textbf{length}}} & min. & $1$ \\
    && max. & $10$ \\ \cline{2-4}
    
    & \multirow{2}{*}{\textbf{training}} & batch size & $16$ \\
    && epochs & $20$ \\ \hline
    
    \end{tabular}
    \caption{Experiment hyperparameters. \\
    $d_{model}$: embedding/hidden layer dimensionality  \\
    $d_{kv}$: key/value matrix dimensionality  \\
    $d_{ff}$: output dimensionality of feed-forward layers}
    \label{tab:hyperparameters}
  \end{center}
\end{table}


%
%
\section{Training, evaluation, and test results}
\label{sec:appendix-results}

Table \ref{tab:results-all} shows training, evaluation, and test results across all fine-tuning tasks in both zero-shot and vocabulary flip settings.
Accuracy is used as the performance metric, and task classes are always divided $50\%-50\%$.

\begin{table*}[t]
  \begin{center}
    \begin{tabular}{|c|c|c|c|c|c|c|} \hline
    \multirow{2}{*}{\textbf{Task}} & \multirow{2}{*}{\textbf{Task class}} &
    \multirow{2}{*}{\textbf{Dataset}} &
    \multirow{2}{*}{\textbf{Zero-shot}} & \multicolumn{3}{c|}{\textbf{Vocabulary flip}} \\ \cline{5-7}
    &&&& \mix$=0\%$ & \mix$=1\%$ & \mix$=10\%$ \\ \hline
    
    \multirow{6}{*}{copy/reverse} & \multirow{3}{*}{copy} & train & $1.00$ & $1.00$ & $1.00$ & $1.00$  \\
    && eval & $1.00$ & $1.00$ & $1.00$ & $1.00$ \\
    && test & $1.00$ & $0.75$ & $1.00$ &
    $1.00$ \\ \cline{2-7}
    & \multirow{3}{*}{reverse} & train & $1.00$ & $1.00$ & $1.00$ & $1.00$ \\
    && eval & $1.00$ & $1.00$ & $1.00$ & $1.00$ \\
    && test & $0.97$ & $0.72$ & $0.99$ & $0.99$ \\ \hline
    
    \multirow{6}{*}{\shortstack{copy/reverse\\detection}} & \multirow{3}{*}{copy} & train & $1.00$ & $1.00$ & $1.00$ & $1.00$ \\
    && eval & $1.00$ & $1.00$ & $1.00$ & $1.00$ \\
    && test & $1.00$ & $0.00$ & $1.00$ & $1.00$ \\ \cline{2-7}
    & \multirow{3}{*}{reverse} & train & $1.00$ & $1.00$ & $1.00$ & $1.00$ \\
    && eval & $1.00$ & $1.00$ & $1.00$ & $1.00$ \\
    && test & $0.88$ & $0.00$ & $0.94$ & $1.00$ \\ \hline
    
    \multirow{6}{*}{\shortstack{palindrome\\detection}} & \multirow{3}{*}{palindrome} & train & $1.00$ & $1.00$ & $0.99$ & $1.00$ \\
    && eval & $1.00$ & $1.00$ & $0.99$ & $1.00$ \\
    && test & $1.00$ & $0.07$ & $0.09$ & $0.90$ \\ \cline{2-7}
    & \multirow{3}{*}{non-palindrome} & train & $0.98$ & $1.00$ & $0.99$ & $0.98$ \\
    && eval & $0.98$ & $1.00$ & $0.99$ & $0.97$ \\
    && test & $0.90$ & $0.00$ & $0.02$ & $0.91$ \\ \hline
    
    \multirow{6}{*}{\shortstack{repetition\\detection}} & \multirow{3}{*}{repetition} & train & $1.00$ & $1.00$ & $0.99$ & $0.93$ \\
    && eval & $1.00$ & $1.00$ & $0.99$ & $0.93$ \\
    && test & $0.96$ & $0.08$ & $0.27$ & $0.30$ \\ \cline{2-7}
    & \multirow{3}{*}{no repetition} & train & $1.00$ & $1.0$ & $0.99$ & $0.91$ \\
    && eval & $1.00$ & $1.00$ & $0.99$ & $0.91$ \\
    && test & $1.00$ & $0.10$ & $0.12$ & $0.10$ \\ \hline
   
    \end{tabular}
    \caption{Training, evaluation, and test accuracy per task class in zero-shot and vocabulary flip settings.}
    \label{tab:results-all}
  \end{center}
\end{table*}

\end{document}